# Title: Using LLMs to create analytical datasets: A case study of reconstructing the historical memory of Colombia

**Authors:** David R. Anderson[1]†, Galia J. Benítez[2]†, Margret V. Bjarnadottir[3]*†, Shriyan Reyya[3]†

**Affiliations:**

[1]Department of Management and Operations, Villanova School of Business, Villanova, PA, 19083

[2]James Madison College, Michigan State University, East Lansing, MI, 48825

[3]Decisions, Operations and Information Technology, University of Maryland, College Park, MD 20742

*Corresponding author. Email: mbjarnad@umd.edu.

† These authors contributed equally to this work

**Abstract:**

Colombia has been submerged in decades of armed conflict, yet until recently, the systematic documentation of violence was not a priority for the Colombian government. This has resulted in a lack of publicly available conflict information and, consequently, a lack of historical accounts. This study contributes to Colombia's historical memory by utilizing GPT, a large language model (LLM), to read and answer questions about over 200,000 violence-related newspaper articles in Spanish. We use the resulting dataset to conduct both descriptive analysis and a study of the relationship between violence and the eradication of coca crops, offering an example of policy analyses that such data can support. Our study demonstrates how LLMs have opened new research opportunities by enabling examinations of large text corpora at a previously infeasible depth.

**Main text:**

**Introduction:**

Colombia has been submerged in decades of armed conflict. In a watershed effort to de-escalate, the government negotiated the 2016 peace agreement with the Revolutionary Armed Forces of Colombia (FARC), the longest-running guerrilla group in the world. This considerably reduced the level of violence in Colombia, and the murder rate has since dropped by over 82% (*1-2*). However, established groups, such as the National Liberation Army (ELN) guerrilla organization, and newer groups, such as FARC dissidents and paramilitary successor groups, are still engaged in forced displacement, killings, kidnappings, disappearances, and massacres. These violent events have increased in recent years, threatening the post-2016 improvements (*2*).

Colombia is the world's largest producer of coca leaf, which is the raw material in the manufacturing of cocaine (*3*). In Colombia, the cultivation and trafficking of cocaine have been inextricably linked with violence, criminal organizations and armed groups (*4-6*). Criminal organizations driven by financial incentives safeguard coca cultivation areas and the lucrative trafficking routes associated with them, often resorting to intimidation and violence. This can involve threats, extortion, and physical attacks against farmers, community leaders, and anyone deemed a threat to the armed groups' operations.

Until the recent peace accord, the systematic documentation and recording of violent incidents and casualties resulting from the armed conflict in Colombia was not an active policy priority for the Colombian government. This neglect can be attributed to a complex interplay of factors, including a pervasive widespread culture of impunity, a deep disregard for human rights, and, in some cases, direct involvement of the government in the conflict. Additionally, challenges in identifying victims and perpetrators, coupled with the significant risks to those denouncing the violations and collecting such data, have hindered governmental and private actors' data collection efforts. These issues have fostered a climate of intimidation and fear among the broader population, allowing perpetrators to operate with relative impunity. As a result, the scarcity of publicly available information about the conflict has contributed to a lack of transparency and comprehensive historical accounts. This lack of historical accounts then significantly hinders policy analysis, particularly in areas such as: evaluating the impact of coca eradication efforts on violence, identifying organized criminal groups and their territorial control, and mapping potential cocaine trafficking routes and associated patterns of violence.

Within the framework of the peace accord, governmental and non-governmental organizations (NGOs) have initiated efforts to collect data. This endeavor, however, is complex, multifaceted, and possesses inherent limitations. In this paper, we therefore explore the potential of large language models (LLMs) as an automated aid to this data collection process.

To do this, we harness over 235,000 news articles about violent incidents. We systematically process and analyze these articles using GPT 4o-mini (see Figure 1 for three example articles), resulting in a transactional dataset of violent events. This dataset enhances the official records of violent events by providing complementary information, extending temporal or geographical scope, offering contextualization, and adding significant depth and granularity. It thus furthers our goal of contributing to the historical memory of Colombia, offering insights into the formation, interpretation, and remembrance of historical events. Practically, the dataset can support descriptive and policy analysis. Methodologically, our study demonstrates how LLMs have opened new research opportunities by enabling examinations of large text corpora at a previously infeasible depth.

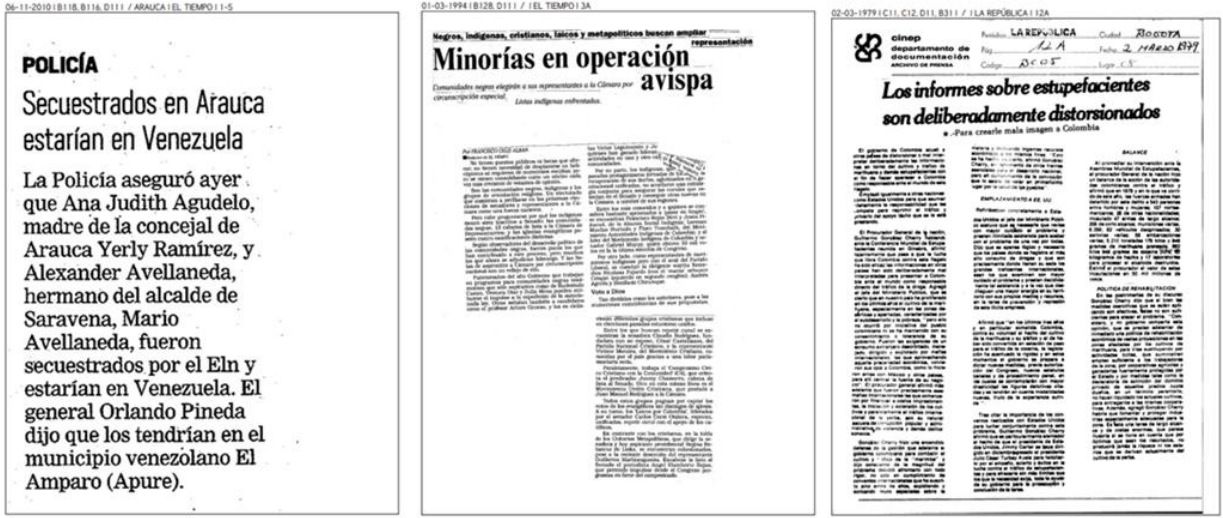

**Fig. 1. Three example articles**. Scans of three newspaper articles used in our analysis.

**Results:**

Our finalized dataset contains 78,685 violent events. Murders are the most frequently reported type, followed by armed conflicts and attacks. The data shows an overall increase in violence between 2005 and 2015, followed by a decline after the peace accord was signed in 2016. We note a relatively low number of less violent crimes (e.g. harassment), as violent crimes are less likely to be reported on by the press.

To contrast the patterns of violence in 2013 and 2019, Figure 2 uses the extracted event locations, the information about who the attackers were (filtered to include FARC, the National Liberation Army (ELN), and the government), and the type of violence. We note that most of the violence in 2013 was committed by FARC and was concentrated in the southwest, particularly in the Cauca Valley. Cauca is a strategic corridor linking the Atlantic and Pacific Oceans, and it has a high concentration of coca cultivation. Consequently, various armed groups fiercely compete to control this vital territory, leading to significant violence and instability. Contrasting the events in 2013 with those in 2019, following the peace process, highlights a dramatic reduction in the amount of violence and a shift in the perpetrators. The violence is now more concentrated in the north, in the states of Santander and North Santander, and is more likely to be committed by the ELN than by FARC. The presence of these groups and associated violence is closely linked to cocaine trafficking and other illicit activities. Therefore, identifying these corridors and groups may help pinpoint areas prone to turf wars and inform government strategies for resource allocation and anti-narcotic policy development.

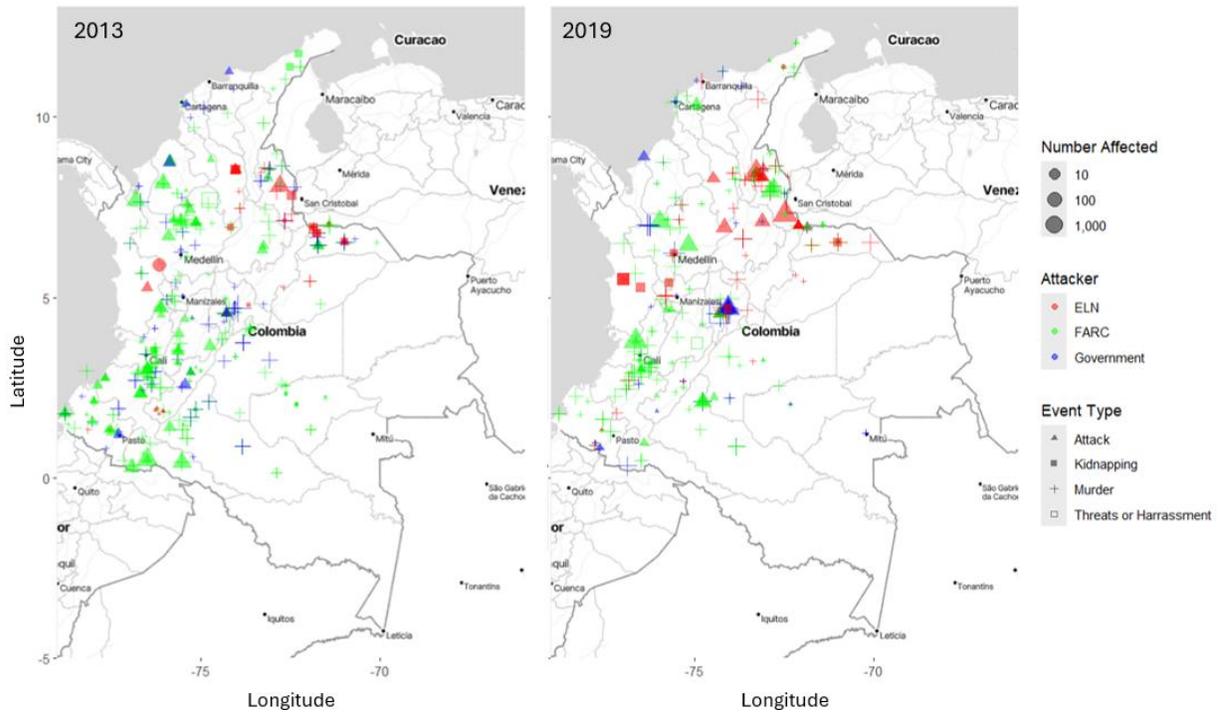

**Fig. 2.** Map of violent events in 2013 and 2019 where FARC, ELN, or the government (including the army) were reported as the "attacker."

Beyond simple descriptive analysis, our dataset offers details and insights far exceeding the limited databases available on violent events in Colombia. As detailed in the Supplement to this paper, we compare our dataset to that maintained by the Center for Historical Memory (CHM) in Colombia. While there are events in the CHM dataset that are not contained in our dataset and vice versa, we compare here the details available for an event contained in both datasets.

Specifically, an event from CHM's Civil Muertos (murders) file is described as occurring on June 20, 2011 in the San Pablo Municipality of the Bolivar Department. There was one victim, the type of killing is classified as "Bombardeo," or bombing, and the party involved is the Fuerza Aerea Colombiana, or Colombian Air Force. Our summary of an article corresponding to the same event is as follows: "The article reports on the death of a 17-year-old boy, Adinson de Jesufas Baquero Valencia, allegedly due to a bombing by the Colombian air force and national police in San Pablo, Bolivar. The authorities are investigating the incident, and preliminary forensic reports indicate the boy's body had four gunshot wounds." The differences in the level of information about this event, and in broader comparisons between the CHM data and our dataset, clearly demonstrate the power of the information gleaned from newspaper articles, which are typically written to convey relevant facts and context. This death, represented only as a single number in the current, publicly available CHM data, encapsulates a 17-year-old boy with four gunshot wounds. The ability to extract such granular details at scale has important implications for policy efforts and for the reconstruction of the historical memory of Colombia.

As an illustrative example of how our data might be used, we zoom in on one city and study the trends of violence in Cali, the capital of the Valle de Cauca Department in Southwest Colombia. Our data reports hundreds of events per year the 2000s and 2010s, with dozens of murders and high-profile killings. These include civilian massacres as well as targeted murders of judicial workers, journalists, and the Archbishop of Cali. For each year from 2006 to 2010, there are over 100 events in our database, with FARC involvement in almost half of the events. After the peace process, the number of events drops significantly, to an average of 38 events per year from 2020 to 2022. FARC involvement becomes minimal, but ELN activity rises from less than 1% of events to over 10%. Armed conflicts have also essentially disappeared, and after 2020 the violence is on a smaller scale, with each event impacting fewer people on average: over half of events involve only one victim, compared to only 25% involving one victim in the 2006 to 2010 period.

Policy Analysis

As an example of the type of policy analysis that our data can support, we address the open question of whether there is a relationship between coca plant eradication efforts and violence. The study of the relationship between coca cultivation and armed conflict is inherently complex. Two prior studies have begun to explicitly explore the connection between forced coca eradication campaigns and violence (*7-8*) without finding conclusive evidence (for details please refer to the Supplement). Our study contributes to this literature by studying a broader range of violence data for a longer time period. In contrast to the previous studies, we find no statistically significant relationship between eradication and violence, whether examining all violent events or specific subsets such as murders or armed conflict (full regression results for all events, murders and armed conflict are included in the Supplement).

**Interpretations:**

LLMs and related technologies have opened up new research opportunities by enabling scholars to conduct in-depth examinations of large text corpora that were previously considered too difficult and time-consuming to analyze (*9-10*). In this project, we have demonstrated how a carefully constructed data mining pipeline (please refer to the Materials and Methods for details), including the use of GPT at scale, can provide new insights into historical events and how subsequent analysis can support policymakers. Our suggested pipeline is not limited to humanities and historical events but can be used in any context where information embedded in text may be useful for analytical tasks.

Some key lessons from our implementation that have broad applicability include the need for very detailed questions in prompts used to featurize datasets. For example, we found that GPT was much more accurate when we asked it whether FARC was involved in an event than when we asked it more generally who the attacker was. We also found that iterative manual reviews of the results were critical in improving the resulting data. Lastly, we observed that lowering the temperature of GPT led to more consistent, accurate, and reproducible results. Typically, the model performed better on yes/no questions than on open-ended questions (e.g. "Was this a murder?" vs. "What type of violent event was described?").

This study is not without limitations, which we acknowledge. While the accuracy of GPT is not perfect, in our evaluations its accuracy is comparable to human parsing of often difficult-to-read images of news articles. We expect that improvements in our prompts, our post-processing (including our deduplication algorithm), and the rapid advancement of LLMs will further improve the accuracy of featurization.

This study is aligned with our ongoing efforts to contribute to Colombia's peace-building efforts by uncovering obscured accounts, illuminating diverse perspectives, and challenging prevailing narratives that have shaped collective memory. By harnessing our extensive reservoir of digital knowledge, we have demonstrated how, with the assistance of GPT, valuable insights into Colombia's human rights violations, violence, and criminal organizations can be obtained, contributing not only to policy decisions but to national collective memory of historical events.

For humanities, social science, and qualitative researchers, LLMs in general and GPT specifically are proving instrumental in expediting the processing of vast quantities of raw digital data with minimal resources and personnel. Tasks that traditionally would have taken months or even years can now be completed efficiently by a small team. Combined with text and data mining capabilities, these developments represent a significant advancement that can empower humanities and social science researchers to synthesize information, identify patterns, and extract key insights that previously would have required large teams. Moreover, GPT's capacity for accurate language translation and processing across multiple languages is a critical asset for researchers who do international work. Its functionality reduces the constraints often associated with assembling teams of translators and language specialists. This streamlines research conducted in diverse linguistic contexts, which commonly introduces significant complexities for researchers. We hope that our study, including our data mining pipeline, can serve as a template for these efforts.

**Supplementary Materials**

Materials and Methods

Supplementary Text

Figs. S1 to S2

Tables S1 to S6

# Supplementary Materials for

## Using LLMs to create analytical datasets: A case study of reconstructing the historical memory of Colombia

David R. Anderson, Galia J. Benítez, Margret V. Bjarnadottir*, Shriyan Reyya
*Corresponding author: mbjarnad@umd.edu

**The PDF file includes:**

Materials and Methods
Supplementary Text
Fig. S1 to S2
Tables S1 to S6

**Other Supplementary Materials for this manuscript include the following:**

None

**Materials and Methods**

The goal of our study is to contribute to the historical memory of Colombia, offering insights into the formation, interpretation, and remembrance of historical events. In order to do this, we needed to translate PDF scans of news articles into a transactional dataset of violent events. We describe the key steps needed below, and we provide examples of a descriptive analysis and a policy study on the impact of coca eradication efforts and violence.

Data

The original data consists of 235,000 scanned Spanish-language articles published in Colombian national and regional newspapers from 1992 to 2022. The articles were collected by the Cinep-Peace program[1], a Colombian non-profit organization that focuses on preserving memories of violence in Colombia. The original data consisted of scans of the articles in JPG or PDF format.

The data mining pipeline

Our goal was to transform the JPG and PDF files into a dataset of violent events. We therefore created the data mining pipeline summarized in Figure S.1. We implemented optical character recognition (OCR) via Google Cloud Vision API to read the scanned images and convert them into text files for processing. After significant prompt engineering (discussed below), the text of the articles was then sent to the ChatGPT 4o-mini (GPT) model via the OpenAI chat-completions API, with the prompt asking GPT to summarize the article and answer a series of questions about the violence reported. The resulting response from GPT was processed into a dataset where each entry corresponds to an article, which was then augmented with longitude and latitude via Google Maps API using the location identified by GPT. As important events may have been covered by multiple media outlets and possibly over multiple days, we developed and implemented a deduplication algorithm to identify multiple reports of the same events. Finally, our data were compared to another available data sources, and quality checks were performed. We discuss the key analytical steps in additional detail below.

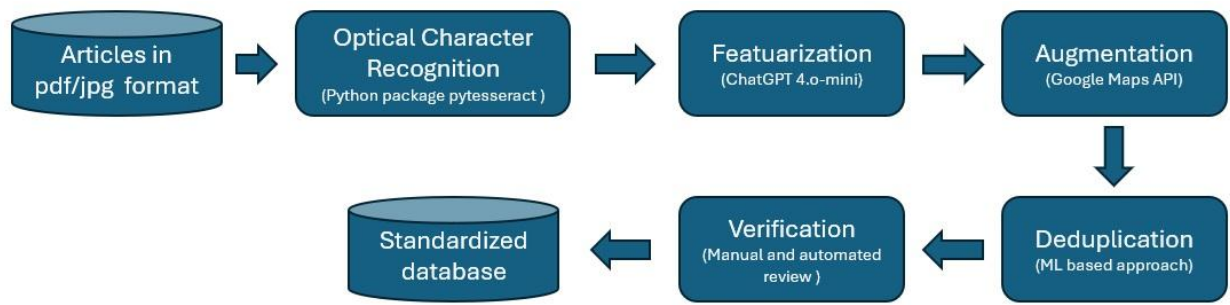

**Fig S.1**: *The key steps in the data mining pipeline.*

Data processing and prompt engineering

Following and expanding on prompt engineering best practices (*12*), we used an iterative process of creating prompts and testing them on 100 randomly selected articles. Each prompt consisted of questions related to the article, for example: "*What words, in Spanish, are used by the author to describe the violence?*" We manually validated the results and scored them for the accuracy of the extracted information and sensitivity (the level of information captured). The questions with low accuracy and/or sensitivity were improved upon in the next iteration by either improving the specificity of the question, splitting up the question, adding additional questions, or giving examples and context.

---

[1] https://cinep.org.co

The finalized prompt used to featurize the articles can be divided into i) instructions and context, ii) 32 specific questions about the articles, iii) formatting guidelines (to ensure a consistent output format, which is critical when featurizing data at scale), iv) request for a summary, and finally v) the article text. The full prompt is provided in Fig. S.2. Each article is then appended to the end of the prompt and the entire prompt and article is passed to the API.

The response to the first question in the prompt ("Is this article primarily about one specific battle, attack or incident of violence? Yes or no?") was used as an inclusion criterion. The articles collected in our dataset are all broadly related to violence in Colombia, but they include discussions of other events including political rallies, memorial events, marches, and protests. Our work is focused specifically on violent events, and so we only include news articles where GPT identifies the article as discussing a specific violent event. This reduced the number of articles to 107,445.

Data quality evaluation and filtering

The largest challenge when extracting event information is working with articles that summarize multiple recent attacks in an area or discuss an emerging pattern of attacks. We found that in such cases, GPT had trouble combining or separating multiple distinct pieces of information. In other words, when an article discusses one attack at one location by one group, with one set of victims, the LLM is nearly perfect at summarizing the information, but when an article discusses multiple events, different victim types, multiple involved parties, or different locations, the model can struggle to parse the correct specific information. Under the assumption that the events reported on in a summary article would have been reported independently, we chose to eliminate articles discussing multiple events. We therefore sent each article summary to GPT and asked whether the article was about one specific event, one focal event and other events mentioned, or multiple distinct events equally. If GPT's answer indicated that the article discussed multiple events, the article was filtered out. This filtered out 59,380 articles that discussed multiple distinct events.

For the remaining articles, we took a number of steps to evaluate the quality of the extracted information. First, we conducted a manual review of 514 randomly selected articles. We found that the location was correct in 507 (98.6% of) cases. Location errors occurred when multiple locations were mentioned in an article. GPT was similarly effective at identifying which group was responsible for the attacks, correctly identifying the group in 489 (96% of) cases. The cases when GPT made errors were when there was ambiguity around who the attackers were. We also found (per the discussion on prompt engineering above) that splitting up the attacker question improved the information accuracy in cases when attackers were not correctly identified. For example, asking directly about FARC's involvement correctly identified FARC when the general question about the attacker did not return an answer. In our manual review, we found 18 (3.5%) articles where GPT either misidentified the group or did not correctly detect the groups involved in the event. The highest number of errors was found in the question focused on the number of victims, with GPT making 53 errors in reporting on 514 articles, or 10.3%. These errors usually occurred when the article mentioned multiple sets of victims or different types of victims. For example, one article reported 14 deaths and 23 injured, but GPT reported 14 victims. Data quality review results are reported in Table S.2.

The last quality concern pertains to OCR failure. In a small number of cases (4 out of the 514 reviewed articles), the OCR process rendered the original article unreadable, but the LLM still answered the questions, "hallucinating" the answers. We addressed this issue by pre-screening the articles and excluding articles shorter than 500 characters, which were consistently subject to OCR failure.

Geocoding

To analyze the events' spatial dimension, we translated the locations identified by GPT into latitude and longitude coordinates that can be used to generate maps or used in spatio-temporal analyses of violent events. Specifically, we used the Google Maps API through the ggmap R package (*13*) to map each location to specific latitude and longitude coordinates. Each location was concatenated with " Colombia" and then encoded with ggmap.

Data deduplication

In the original data, certain events were covered multiple times, either because each event was discussed by multiple news sources and/or because the event was covered over the course of multiple days. In similar situations, previous studies have implemented several approaches for deduplication of large text and event corpora. The shared basis for most approaches involves embedding texts (*14*) and computing similarities with various similarity metrics. Additional features like location and date are incorporated when available, either directly as a requirement for a match (e.g. the event needs to take place within a specific time window (15) or weighted by value in the final classification. The different elements are either combined manually into a similarity score (*15-17*), or a classifier can be built if labeled data is available (*17-18*). For example, Piskorski et al. explored various text similarity metrics and found that combining text and metadata features significantly improved performance (*19*). More recently, authors have explored the use of neural networks as more advanced machine learning approaches to improve deduplication, for example to address noise issues (e.g. OCR errors) (*18, 20*).

We adopt and build on the published literature to deduplicate our data. Given the share size and relatively large dimensions of the data, we implemented a supervised machine learning approach. As there is no "ground truth," we first selected four representative three-month periods, with a total of 2,245 articles, and manually labeled multiple observations describing the same event. The manual labeling process resulted in 734 articles (24.2%) identified as duplicates describing 189 events, with the remaining articles each discussing a unique event.

We next implemented a machine learning approach on the labeled data. For each pair of articles in the labeled data, and focusing on densely populated columns, we created variables indicating whether the information in the two articles matched or was similar. For example, for the date column, we calculated the number of days between two event dates. For binary columns, namely those describing the type of conflict (e.g. murder, armed conflict, kidnapping) or a group (e.g. FARC, ELN, EPL), we created a three-level categorical variable (indicating whether both values were 1, both 0, or they do not match). For the locations mentioned and the key Spanish words, we created a binary indicator to denote whether there exists an intersection between the two records, and another binary indicator to denote whether the first location mentioned was the same. Finally, for both the article text and article summary (provided by GPT), we used text cosine similarity score to quantify the similarity between two texts. This involved shingling (breaking the text into sequences of overlapping tokens) each text into n-grams (with the value of n set to 2) for contextual embedding, vectorizing with term frequency-inverse document frequency on the union between each set of shingles, and finally calculating the cosine similarity between the two text vectors. The full list of variables used in modeling is provided in Table S.1.

We split the labeled data into training and testing, creating a testing dataset with 20% of events, none of which overlapped with the remaining 80% in the training dataset. We then trained multiple models: Logistic Regression (*21*), Support Vector Machine (*22*), Multilayer Perceptron (*23*), Random Forest (*24*), and XGBoost (*25*). The Random Forest model had the best performance on a balanced training dataset where we down-sampled non-duplicate pairs. The feature importance from the final Random Forest Model (100 trees in forest, trained on bootstrap data, no max tree depth) is provided in Table. S.1. The most important model feature was the original Spanish article text cosine similarity, followed by locations (whether the list of locations had an overlap and whether the first mentioned location was the same), and the number of days between the articles' publication dates.

The resulting model has an 83% overall accuracy, 95% sensitivity, 72% specificity, 83% AUC, and 91% F-measure all measured on a balanced test set, metrics which are in line with the literature. Importantly, we note that the difficulty level of deduplication is very dataset dependent, and therefore direct comparisons may not be meaningful. With that in mind, we note that Zavarella et al. reported an F-measure of 91.5% (*16*), similar to ours, and Piskorski et al., when including metadata features in addition to text similarities, reported F-measures of 98.6% and 97% on two different datasets (*19*). Rollo and Po reported an accuracy between 93 and 94% and an F-measure between 71% and 73% (*15*).

We next applied the model to the remaining dataset, comparing events dated within one month of each other. We set the duplicate classification model cutoff threshold as 0.95, that yields an overall duplicate rate close to the rate estimated from manual labeling. We used a graph-clique-identification approach (*17*), constructing a graph in which articles represented nodes and edges were weighted by the model classification probability. In lieu of assuming transitivity across article chains, each article in each identified duplicate group was classified as describing the same event as every other article in the group (i.e., only complete sub-graphs, or cliques, were targeted), mapping each record to a specific event, and keeping the article with the most information, based on an information score in which location weighed twice as much as other factors, from each event. After removal, 78,685 events were retained, with 65,178 occurring between 2000 and 2022; this corresponds to removing 27% of the articles as duplicates.

Comparison with other data sources

To understand the value of the extracted data, we compared our data with a dataset collected by the Center for Historical Memory (CHM) (https://centrodememoriahistorica.gov.co/). CHM's aim is to collect testimony and preserve the collective memory of armed violence in Colombia (*27*) and has put together the Information System on Violence in the Colombian Armed Conflict (Sistema de Información de Eventos de Violencia del Conflicto Armado Colombiano - Sievcac), a comprehensive database that systematically collects and consolidates quantitative data from numerous sources monitoring the nation's internal conflict. A key feature of this information system is its commitment to data integrity, achieved through a permanent updating process and a meticulous case-by-case deputation and cleaning protocol. This rigorous methodology establishes Sievcac as the most significant and robust database of its nature. The events in the CHM dataset span from 1970 to 2012 and are organized into categories of kidnappings, killings, deaths, massacres, attacks, mines, and terrorist attacks. We excluded mines from the comparison.

To identify overlap and differences between the two datasets, we compared every entry in our dataset to all events occurring in the same or adjacent months of each entry in the CHM dataset where key characteristics match. In particular, we required the following characteristics to match: i) the date needed to be in the same month or adjacent months (our data is at the month level), ii) at least some (upper bound on matches) or all (lower bound on matches) of the parties involved needed to match, and iii) the location (longitude and latitude in our dataset mapped into municipios reported in CHM's data) needed to be at most 20km (lower bound) or 40km (upper bound) apart. Additionally, the violence types needed to match.

Our comparison found that between 5.9% and 10.2% of events in our dataset were also reported in the CHM dataset, and these events comprised between 11.5% and 19.8% of CHM's total data. To further gain insights into differences between the datasets, we randomly selected events that appeared in both datasets, exclusively in ours, and exclusively in CHM data, and studied the differences (we provide detailed examples of matches and non-matches in the supplementary text). The comparison reveals that our data provides detailed insight into the events themselves, including political motivations, attack methodologies, and more granular information about the groups involved. Our dataset also reflects the tone and emotion of the community, as communicated in the articles.

Policy analysis

To analyze the relationship between coca crop eradication and violence, as motivated in the main body of the paper, we used data about areas affected by coca cultivation and anti-narcotic policies (aerial spraying and manual eradication) collected from the Integrated System for Monitoring Illicit Crops (*Sistema Integrado de Monitoreo de Cultivos Ilícitos*) (*28*), which includes medium-resolution satellite images and field surveys of coca plantations conducted by the United Nations Office on Drugs and Crime. Eradication is measured in hectares of affected drug crops at the municipality level from 1999 to 2022. Municipality-level eradication records were then assigned to the department in which the municipality is located.

We ran a fixed effects regression analysis controlling for the number of violent events in previous years and overall coca eradication, both at the department level. Based on the statistical distributions of our data we took the natural log of both the hectares of eradication (plus 1) and number of violent events (plus 1), in each geographical area, each year. The fixed effects control for the average level of violence in each year and in each department, meaning that the regression coefficients now measure the localized impacts of more violence in the preceding year(s) and the association with more eradication in the years immediately prior. More specifically, we estimate the following regression for the log of the number of events in location $k$ in year $t$:

$$\text{Ln (Number of Events}_{tk}+1) = \alpha + \Sigma_{i=t-3}^{t-1} \beta_{ei} \ln(\text{Number of Events}_{ik}+1)$$
$$+ \Sigma_{i=t-3}^{t-1} \beta_{hi} \ln(\text{Hectares Eradicated}_{ik}) + \beta_k \text{departmento}_k + \beta_t \text{year}_t + \varepsilon$$

In addition to this model specification, we tested multiple other model specifications as robustness checks. These checks include using linear and logged dependent and independent variables, using one to five lagged treatment variables, using binary (any eradication), linear, and logged treatment effects (hectares eradicated, ln(hectares eradicated)), and testing models with manual versus aerial eradication as separate variables. In no case were we able to detect any statistically significant effects.

**Acknowledgments:**

**Funding:** This work was in part supported by National Science Foundation grant 2039862 Collaborative Research: D-ISN: TRACK1: Developing Advanced Analytics Tools for Discovery, Analysis, and Disruption of Narcotic Supply Chains (MVB, GB)

**Author contributions:**

Conceptualization: DA, MVB, GB

Methodology: DA, MVB

Investigation: DA, GB, SR, MVB

Visualization: DA

Funding acquisition: MVB, GB

Writing – original draft: MVB, GB

Writing – review & editing: MVB, SR, DA, GB

**Competing interests:** Authors declare that they have no competing interests.

**Data and materials availability:** The code and event summaries will be made available upon the publication of the paper via their GitHub depositories.

**Supplementary Text**

Detailed event comparison

We compared our dataset to a currently available source, that of the Center for Historical Memory (CHM), as detailed in the Materials and Methods section. Below we provide three examples of events found in both datasets, three events found only in our data, and three events found exclusively in the CHM dataset.

Events found in both datasets:

- From CHM's Ataques file: An event that occurred on August 5, 2000, involving the Guerilla-FARC Group, in the Cabecera Municipal of the El Carmen de Atrato Municipality, resulting in 11 victims. We identified this same event in our dataset, and the summary that GPT generated for the matching article is as follows: "The article reports on a violent attack by the farc guerrilla group in carmen de atrato, resulting in the deaths of 11 members of the fuerza p[ú]blica. The attack also caused significant infrastructure damage, including the destruction of a police station, part of a church, and a bank. Additionally, the article mentions other violent incidents and attacks by guerrilla groups in various locations." The date information matches, but our data provides more detailed information about the front involved (Frente 34) and specific details regarding the attack, namely relating to infrastructure, some of which is mentioned in the summary.

- From CHM's Asesinatos file: The event occurred on August 18, 2004 in the Tulua Department of the Valle del Cauca Municipality, and various paramilitary groups were involved. There was one victim. Our summary: "The article discusses the murder of campesino leader carlos ovidio agudelo, who was killed with a chainsaw, allegedly by the auc. it also mentions fernando le[ó]n soto berrio, a spokesperson for the auc, who has a history of legal issues and is currently facing multiple charges." The AUC is a prevalent paramilitary group which our data specifically mentions, and the summary also provides the politically relevant information of who specifically was murdered and how.

- From CHM's Civil Muertos file: The event occurred on June 20, 2011 in the San Pablo Municipality of the Bolivar Department. The type of killing is classified as "Bombardeo," or bombing, and the party involved is the Fuerza Aerea Colombiana, or Colombian Air Force. There was one victim. Our summary: "The article reports on the death of a 17-year-old boy, adinson de jes[ú]s baquero valencia, allegedly due to a bombing by the colombian air force and national police in san pablo, bol[í]var. the authorities are investigating the incident, and preliminary forensic reports indicate the boy's body had four gunshot wounds." As with the other matches, here we can see the power of gleaning information from newspaper articles specifically designed to convey relevant facts. This death, represented only as a single number in the current, publicly available data, encapsulates a 17-year-old boy with four gunshot wounds. The ability to extract such granular details at scale has significant implications the reconstruction of the historical memory of Colombia.

The following are examples of events in the CHM dataset that were not identified in in our dataset.

- From CHM's Secuestros file: The event occurred on December 29, 2008 in the Villa Garzon Municipality of the Putumayo Department. A single person was kidnapped, and was likely motivated by economic reasons. The alleged perpetrator is a BACRIM, or banded criminal gang.
- From CHM's Terroristas file: The event occurred on April 15, 2002 in the Jamundi Municipality of the Valle de Cauca Department. There were 2 injuries and 3 deaths, and the terrorist attack was carried out by a non-descript Paramilitary Group.
- From CHM's Ataques file: In the Algeciras Municipality of the Huila Department, explosive artifacts were used by a FARC guerrilla group to kill an unknown number of people.

Finally, the following are examples of event summaries from our dataset that were not found in the CHM dataset.

- "The article reports the discovery of a mass grave in tolima, colombia, containing twelve bodies showing signs of torture and execution. It highlights the involvement of paramilitary and military forces, and mentions ongoing threats and extortion by the auc."
- "The article discusses the tragic story of don leonel granados, whose son was kidnapped and murdered by the farc. his wife was also kidnapped when she tried to negotiate for their son's remains. the article highlights the ongoing violence and extortion faced by civilians in colombia."
- "The article discusses the sexual violence against a 13-year-old indigenous girl in tame, arauca, allegedly by five soldiers. it highlights the ongoing issue of sexual abuse by members of the fuerza p[ú]blica in the region and mentions previous incidents, including a case involving a sub-lieutenant from the brigada m[o]vil 5. the article also notes the community's concern and the ongoing investigation by authorities."

Detailed policy results and comparison with the published literature

There are two earlier papers that study the relationship between forced eradication of coca and violence. The first study used monthly data at the department level to study the impact of eradication on violence levels in the years 2004-2005. It found that an increase in eradication led to a decrease in total violence, and that it was only FARC (and not ELN) showing a reduction in total violence. However, the study was somewhat limited due to its small data sample, with fewer than 700 violent events (203 assassinations, 77 incidents of terrorism, and 404 kidnappings) over the 792 department-month observations. It was also limited in time, lasting only two years, which impeded its ability to measure long-term effects. Rozo studied the impacts of eradication on a range of outcomes and found an increase in violence when aerial eradication increased in the area (*9*). Using officially reported municipality-level homicide rates, the paper reports a 4% increase in homicide rate for every additional 1% increase in area sprayed, and a similar increase in forced displacement. Their paper, however, relies on officially reported statistics, which may under-report the total amount of violence.

We studied a total of 65,178 violent events that took place during the period 2000 to 2022 and are classified as murder, attacks, armed conflict, kidnapping, or threats and harassment. The outcome used in the analysis was the overall number of events, or events of each violence type, in each department each year. We ran three regression analyses to study the relationship between coca eradication and violence, first for all violent events, then for murders only and armed conflict only. Each time, we controlled for the number of violent events and the number of hectares eradicated in each of the previous three years. The results of the regressions are summarized in Tables S.3, S.4 and S.5. We note that there are no statistically significant relationships except for the amount of violence in the department in the prior year, and the positive regression coefficient indicates a positive relationship. Specifically, there is no statistically significant relationship between coca eradication and the total violence in a given department. As discussed in the Materials and Methods section, we ran additional robustness checks across the different outcomes; in no case were we able to detect statistically significant effects.

**Table. S1.**

A list of variables used in the deduplication algorithm and the corresponding random forest variable importance (impurity decrease)..

| Feature | Feature description | Importance |
| --- | --- | --- |
| article_text_sim | TF-IDF weighted similarity score between article texts. | 36.86% |
| first_location_equal | Binary indicator of whether the first-mentioned location in each article matched. | 13.71% |
| num_days_apart | Number of days apart between event dates. | 12.42% |
| locations_equal | Binary indicator of whether there was existing intersection of at least one location between identified locations. | 11.99% |
| dist_apart | Distance (km) apart between coordinate locations of events in each article. | 6.84% |
| kidnapping_status | Binary indicator of whether both articles included a kidnapping. | 3.20% |
| violence_equal | Binary indicator of whether both articles described the same violence type. | 2.76% |
| army_status | Binary indicator of army involvement status match. | 2.13% |
| words_spanish_equal | Binary indicator of intersection of at least one word between Spanish words identified to describe the violence in the article. | 2.00% |
| summary_sim | TF-IDF weighted similarity score between article summaries. | 1.28% |
| murder_status | Binary indicator of whether both articles described a murder. | 1.11% |
| farc_status | Binary indicator of FARC involvement status match. | 1.08% |
| guerrilla_status | Binary indicator of Guerrilla involvement status match. | 1.05% |
| attack_or_injury_status | Binary indicator of whether both articles described attack or injury. | 0.82% |
| auc_status | Binary indicator of AUC involvement status match. | 0.74% |
| armed_conflict_status | Binary indicator of whether both articles described armed conflict. | 0.72% |
| eln_status | Binary indicator of whether both articles described ELN involvement. | 0.61% |
| harassment_or_threats_status | Binary indicator of whether both articles described harassment or threats. | 0.55% |
| epl_status | Binary indicator of whether both articles described EPL involvement. | 0.13% |

**Fig. S.2.** The finalized prompt. Notably, in order to not "hint" to the LLM the desired answer, we do not give example answers for the questions in the prompt, but we specify the requested format in general terms.

**(i) context and instructions**

"Please answer the following questions about the news article below.

For each question if you do not know the answer, report -1.

**(ii) content questions**

A: Is this article primarily about one specific battle, attack, or incident of violence? yes or no?
B: What words, in Spanish, are used by the author to describe the violence? Give up to 10 words.
C: Does it say how many victims there were? If yes, tell me how many.
D: What is the gender of the attackers?
E: What is the gender of the victims?
F: Does the article describe one or more murders?
G: Does the article describe one or more people who were attacked and injured?
H: Does the article describe one or more kidnappings?
I: Does the article describe armed conflict?
J: Does the article describe harrassment or threats of violence?
K: How many of the victims were children?
L: What words, in Spanish, are used by the victims or witnesses to describe the violence?
M: Does this article reference specific locations?
If yes, list up to two locations that were most important.
N: Who were the attackers?
O: Who was attacked? Options can be like civillians, military, guerrila groups, drug traffickers, etc.
P: What month and year did the attack or event occur?
Q: Does it reference dead people or corpses? If so, how many?
R: Was the army a direct combatant in the conflict?
S: Does the article mention guerrilla groups?
T: Was FARC involved?
U: Was AUC involved?
V: Was ELN involved?
W: When was the article published? The first line of the article gives the date the article was published in MM-DD-YYYY format.
X: What is the emotional tone of the article?
Y: What front (frente) or comission (comisión) was involved in the attack?
Z: What block (bloque) or narcoparamilitary group was involved in the attack?
AA: Was the epl (Ejército Popular de Liberación) involved in the attack?
AB: What group names are mentioned in the article related to the attack?
AC: Does the article reference civillians killed by the army? If so, how many?
AD: Does the article reference falsos positivos deaths? If so, how many?
AE: Does the article reference the name of the attacker? If so who?

AF: Does the article reference the name of a criminal group that the attackers belonged to? If so, what is the name of the group?

**(iii) formatting instructions**

Output your response in a semi-colon separated format, with the answer to each question separated by a semicolon.
An example would be:
A: Yes/No; B: word1, word2, word3; C: #/No; D:Gender/-1; E: Gender/-1; F: Yes/No; G: Yes/No; H: Yes/No; I: Yes/No; J: Yes/No; K: #/No; L: -1/word1, word2, word3; M: -1/Location1, Location2; N:-1/ Attacker1, Attacker2; O:-1/ Victim1, Victim2; P: -1/Month Year; Q: #/No; R: Yes/No; S: Yes/No; T: Yes/No; U: Yes/No; V: Yes/No; W: MM-DD-YYYY; X: Tone; Y: Front/Comission; Z: -1/Block/Narcoparamilitary group; AA: Yes/No; AB: -1/Group1, Group2; AC:#/No; AD: #/No; AE: Attacker's Name/-1; AF: Group Name/-1.

**(iv) summary request**

Then give a brief 1-3 sentence summary of the article."

**Table S.2.**
Results of data extraction evaluation based on 514 randomly selected articles.

| Variable | Description |
|---|---|
| Number of Victims | 89.7% Correct |
| Location | 98.6% Correct |
| Attacker Group | 96.0% Correct |
| Victim Type | 84.6% Correct |
| Violence Type | 86.6% Correct |
| Any Violence | 90.2% Correct |

**Table S.3.**

Regression results for all violent events[a].

| Variable | Estimate | Std. Error | t-value | p-value |
|---|---|---|---|---|
| ln(Number of Events +1)t -1 | 0.254 | 0.062 | 4.12 | 0.001 |
| ln(Number hectares treated +1)t -1 | 0.023 | 0.011 | 1.985 | 0.071 |
| ln(Number of Events +1)t -2 | -0.046 | 0.058 | -0.793 | 0.443 |
| ln(Number hectares treated +1)t -2 | -0.017 | 0.014 | -1.186 | 0.258 |
| ln(Number of Events +1)t -3 | 0.016 | 0.054 | 0.299 | 0.77 |
| ln(Number hectares treated +1)t -3 | -0.002 | 0.016 | -0.094 | 0.927 |

[a]Year and department fixed effects omitted for brevity

**Table S.4.**

Regression results for murders[a].

| Variable | Estimate | Std. Error | t-value | p-value |
|---|---|---|---|---|
| ln(Number of Events +1)t -1 | 0.13 | 0.065 | 2.009 | 0.068 |
| ln(Number hectares treated +1)t -1 | 0.027 | 0.022 | 1.257 | 0.233 |
| ln(Number of Events +1)t -2 | 0.027 | 0.057 | 0.478 | 0.641 |
| ln(Number hectares treated +1)t -2 | -0.022 | 0.019 | -1.188 | 0.258 |
| ln(Number of Events +1)t -3 | 0.034 | 0.06 | 0.573 | 0.577 |
| ln(Number hectares treated +1)t -3 | 0.001 | 0.019 | 0.059 | 0.954 |

[a]Year and Department fixed effects omitted for brevity

**Table S.5.**

Regression results for armed conflict[a].

| Variable | Estimate | Std. Error | t-value | p-value |
|---|---|---|---|---|
| ln(Number of Events +1)t -1 | 0.13 | 0.065 | 2.009 | 0.068 |
| ln(Number hectares treated +1)t -1 | 0.027 | 0.022 | 1.257 | 0.233 |
| ln(Number of Events +1)t -2 | 0.027 | 0.057 | 0.478 | 0.641 |
| ln(Number hectares treated +1)t -2 | -0.022 | 0.019 | -1.188 | 0.258 |
| ln(Number of Events +1)t -3 | 0.034 | 0.06 | 0.573 | 0.577 |
| ln(Number hectares treated +1)t -3 | 0.001 | 0.019 | 0.059 | 0.954 |

[a]Year and Department fixed effects omitted for brevity